\documentclass[aps,prl,twocolumn]{revtex4-2}

\usepackage{graphicx}
\usepackage{amsmath,amssymb,amsfonts}
\usepackage{amsthm}
\usepackage{mathrsfs}

\usepackage{booktabs}
\usepackage{listings}
\usepackage{algorithm}
\usepackage{algpseudocode}

\usepackage{xcolor}
\usepackage[hidelinks]{hyperref} 

\begin{document}

\title{Use of a genetic algorithm in university scheduling for equitable and efficient determination of teaching assignments}

\author{Tom Bensky}
\email{tbensky@calpoly.edu}
\affiliation{Department of Physics, California Polytechnic State University, San Luis Obispo, CA, USA}

\author{Karl Saunders}
\affiliation{Department of Physics, California Polytechnic State University, San Luis Obispo, CA, USA}


\begin{abstract}
Here a genetic algorithm (GA) is presented that creates a teaching schedule for a university physics department by algorithmically assigning ${\sim}200$ classes to ${\sim}50$ professors for each of three academic terms per year.  The algorithm is driven by chromosomes of the GA that encode proposed pairings between enumerated lists of professors and classes. The fitness of the pairings is measured by considering both contractual work constraints and individual teaching preferences. The algorithm  uses standard crossover and mutation operations to seek ever more optimal schedules over many generations.  Here we detail the implementation and performance of the algorithm, including some interpretability findings. Overall, we are very pleased with the algorithm, as it is typically able to converge within minutes, with over $90\%$ of needed classes assigned. A metric is used to assign each professor's schedule a score, which measures how well their preferences were satisfied.  These scores can be used to ensure longitudinal equity in the assignment of classes among professors.
\end{abstract}

\maketitle

\section{Introduction}\label{sec1}

\subsection{Observations of a scheduler}\label{Observations of a scheduler}


Central to the mission of any university is the delivery of instruction to its students. In our department, each of three academic terms per year require the assignment of ${\sim}200$ classes to a corresponding group of ${\sim}50$ professors. At our university, the scheduling of classes and assignment of professors is typically done at the department level. Within a large department such as ours, the work is not only logistically difficult, but involves the challenges of ensuring fairness, equity, and transparency among the professors involved.

The assignment of classes is typically done manually\cite{manual} which is both arduous and time-consuming. We find the excellent introduction by Sarin\cite{sarin2009} to reflect the scope, challenges and past efforts for this kind of work.  Further, assigning each professor classes in a serial manner (typical of manual scheduling), is inherently unfair as the number of constraints on downstream professors grow with each completed schedule.  Thus, professors scheduled earlier in the process typically receive ``better" schedules. 

Moreover, it is not clear how to measure the quality of a given professor's schedule so descriptions such as ``better'' or ``worse'' are rather arbitrary. This is unfortunate since any resulting schedule is highly visible to its participants, who may see any ``best effort'' as inequitable\cite{muhlenthaler2013}, which in turn can lead to significant and widespread job dissatisfaction. 
Thus, it is in the interest of a university to strive 
to 
develop the best possible scheduling workflows and outcomes. 

In this article we present a computer-automated method of assigning classes, driven by a genetic algorithm, that we believe is a significant improvement in terms of efficiency, equity, and transparency (not to mention the enormous time savings). This approach was designed and implemented by the department scheduler (TB) and chairperson (KS) to assign classes in our large physics department\cite{calpoly} over many academic terms. 

\subsection{Paper outline}

Below, we begin by discussing our initial thoughts on developing a computer-automated scheduling process that works with our particular scheduling needs, and how searching for an optimal schedule with a GA in particular seems to cover all requirements we seek. This is followed by a thorough description of our scheduling problem, and how we developed the fitness function for the GA. Next, we present how the evolving chromosomes are parsed to make class assignments, then present some interpretability and explainability findings resulting from use this algorithm for an actual academic term.  Some final optimizations of the schedule using the Prolog language are then presented.

\section{Considering computer automated scheduling assistance}

Existing routes to computer assisted scheduling\cite{burke2002,burke2004d,mirhassani2006,tripathy1984,dimopoulou2001,oss} both open source and commercial\cite{oss,commercial} were investigated for our needs, but nothing suitable was found.  Projects on Github\cite{github} were also a potential resource, but many were found to be incomplete, with little testing or documentation, and were mostly small experiments in scheduling or thesis projects not ready for practical implementation\cite{edceliz}.  Even if a suitable existing solution was found, adapting it for {\sl our scheduling needs} would still be difficult.

The Prolog language\cite{clocksin} has a reputation for being able to find solutions to complex problems like scheduling\cite{prolog_schedule}. However whether using Prolog or some other constrained problem solver\cite{minizinc}, we were not prepared to programmatically handle both hard and soft scheduling constraints\cite{hard_soft}.  Realizing that scheduling could be enumerated over the integers\cite{daskalaki2004,deris1997}, we considered constraint logic programming over the finite domain in Prolog\cite{triska_clpfd}, which has a convenient library\cite{clpfd}. Such a library endows Prolog searches with a range of solution acceptability, somewhat relaxing the strict true or false answers typical of Prolog. There is even a timetabling example built with such\cite{triska}, but we were unable to make adequate progress using these tools (but do see their potential).

\subsection{Settling on a Genetic Algorithm}

From other work in physics (unrelated to scheduling), we are aware of optimization techniques such as the Levenberg-Marquardt algorithm\cite{lm} and others\cite{nr}, including ``genetic algorithms'' (GA)\cite{koza,michalewicz}. It was however, difficult for us to see how the traditional numerical techniques could be used for in scheduling, in for example imagining a ``function'' to optimize, possible mathematical derivatives for it, or even what the domain space would be.  In our view, the GAs have some ``non-numeric'' appeal, given the apparent simplicity of its core premises, which are (initially) random chromosomes (here binary strings of ones and zeros), followed only by the actions of crossover and mutation\cite{koza,michalewicz} to drive the optimization. 

Additionally, in planning a computer-automated scheduling algorithm, we demand three key actions the algorithm must deliver, which we think of as the ``three As'' of scheduling:  1) \underline{a}ssigning classes to professors, 2) \underline{a}ssessing the quality of a given schedule and 3) \underline{a}dapting the schedule repeatedly until we are satisfied with its results.  As we will show, a GA is naturally able to accommodate all of three of these.

\section{Scheduling data and initial constraints}

\subsection{Our basic scheduling problem}

Within our university system, faculty workload is measured in ``units'' where a full academic term workload is 15 units. The fraction of this workload dedicated to teaching is referred to as ``teaching workload.'' The teaching workload for each professor varies between 0 and 15 units\cite{non teaching workload}, as shown in Table~\ref{prof_capacity}.
\begin{table}[h]
\centering
\begin{tabular}{c|c}
\toprule
Professor & Teaching Workload Units\\
\midrule
Prof$_1$   & 12\\
Prof$_2$    & 7\\
Prof$_3$    & 15\\
Prof$_4$ & 2\\
$\vdots$ & $\vdots$\\
Prof$_N$ & Units$_N$\\
\hline
\end{tabular}
\caption{Teaching workload units for each professor. Within the California State University system, each professor has a workload of 15 units per academic term, a fraction of which is dedicated to teaching\cite{non teaching workload}. 
}
\label{prof_capacity}
\end{table}

The courses offered by our  physics department can be generally be classified as ``major," ``non-major," and ``speciality." ``Major" courses are those that we offer to students who are completing a degree in physics. Each term ${\sim}10$ courses are scheduled for major students, which are manually assigned by the department scheduler and chairperson\cite{Major Courses}. There are also several ``speciality" geology and astronomy courses offered each term, for which professors are also selected by the scheduler and chairperson. The vast majority (90$\%$) of courses offered by the department however, are for non-major students from across the university for whom a physics course is part of their degree requirement. This would include for example, students completing engineering degrees. 

The first step of the scheduling process is to assign a professor to each of the major and speciality courses, each of which has its own meeting pattern (days and times). The remaining (and by far the most challenging) part of the scheduling process is to assign the remaining non-major courses. There are six flavors of these courses, four of which have both lecture and lab components, and two of which that just have lecture components. These courses are listed in Table~\ref{non-major courses}. 

\begin{table}[h]
\centering
\begin{tabular}{c|l}
Non-major Course & Modes\\
\midrule
PHYS 121 & Lecture Only \\
PHYS 122 & Lecture and Laboratory \\
PHYS 123 & Lecture and Laboratory \\
PHYS 141 & Lecture Only \\
PHYS 142 & Lecture and Laboratory \\
PHYS 143 & Lecture and Laboratory \\
\end{tabular}
\caption{A list of non-major courses along with the modes of instruction for each.}\label{non-major courses}
\end{table}

Each course has many sections in order to accommodate the large number of students that need to take the course, typically ${\sim}100$ lecture and laboratory sections of non-major courses per term. Indeed, the vast majority of our department work capacity (and teaching units) is devoted to teaching these sections. At the stage in our scheduling process when we need the GA, each lecture and laboratory section have already been placed in a room with its appropriate time pattern\cite{room_optimize}. Thus our only concern is only with assigning professors to these classes. Table~\ref{classes_offered} shows a list of the non-major sections along with the teaching workload and meeting pattern for each section.

\begin{table*}[t]
\centering
\begin{tabular}{c|c|c|c|c|c}
\toprule
Course & Section &  Workload Units & Mode  & Meeting Days & Meeting Times\\
\midrule
PHYS 121 & 01  & 4 & Lecture & Mon, Tue, Wed, Thu & 9am-10am\\
PHYS 121 & 02  & 4 & Lecture & Mon, Tue, Wed, Thu & 10am-11am\\
$\vdots$ & $\vdots$ & $\vdots$ & $\vdots$ & $\vdots$ & $\vdots$\\
PHYS 121 & 7  & 4 & Lecture & Tue, Thu & 2pm-4pm\\
PHYS 122 & 01  & 3 & Lecture & Mon, Wed, Fri & 8am-9am\\
PHYS 122 & 02  & 2 & Laboratory & Wed & 12pm-3pm\\
PHYS 122 & 03  & 2 & Laboratory & Thu & 3pm-6pm\\
$\vdots$ & $\vdots$ & $\vdots$ & $\vdots$ & $\vdots$ & $\vdots$\\
PHYS 122 & 24  & 2 & Laboratory & Fri & 8am-11am\\
PHYS 123 & 01  & 3 & Lecture & Mon, Wed, Fri & 9am-10am\\
$\vdots$ & $\vdots$ & $\vdots$ & $\vdots$ & $\vdots$ & $\vdots$\\
PHYS 141 & 01  & 4 & Lecture & Mon, Wed, Fri & 9am-10am\\
$\vdots$ & $\vdots$ & $\vdots$ & $\vdots$ & $\vdots$ & $\vdots$\\
PHYS 141 & 14  & 4 & Lecture & Tue, Thu & 4pm-6pm\\
$\vdots$ & $\vdots$ & $\vdots$ & $\vdots$ & $\vdots$ & $\vdots$\\
PHYS 142 & 01  & 3 & Lecture & Mon, Wed, Fri & 8am-9am\\
PHYS 142 & 02  & 2 & Laboratory & Wed & 12pm-3pm\\
PHYS 142 & 03  & 2 & Laboratory & Thu & 3pm-6pm\\
$\vdots$ & $\vdots$ & $\vdots$ & $\vdots$ & $\vdots$ & $\vdots$\\
PHYS 143 & 01  & 2 & Laboratory & Fri & 3pm-6pm\\
$\vdots$ & $\vdots$ & $\vdots$ & $\vdots$ & $\vdots$ & $\vdots$\\
PHYS 143 & 36  & 2 & Laboratory & Wed & 3pm-6pm\\
$\vdots$ & $\vdots$ & $\vdots$ & $\vdots$ & $\vdots$ & $\vdots$\\
\hline
\end{tabular}

\caption{A list of courses offered in a given term. The first column is the course name. The second column is the section number. The third column is the teaching workload for that course. The fourth and fifth columns show the meeting days and times for that section.}
\label{classes_offered}
\end{table*}
The basic scheduling problem for us then is to assign the sections listed in Table~\ref{classes_offered} to the professors listed in Table~\ref{prof_capacity}. Professors will (in general) have multiple courses/sections assigned to them. The only {\it constraints} are that there are no time conflicts among those courses assigned to a given professor, and that the total teaching workload does not exceed the mandated teaching workload of that professor. This problem is reminiscent of the ``Knapsack Problem''\cite{knapsack,vahdatpour}.


\section{Departmental and Faculty Preferences}

\subsection{Departmental Preferences}\label{dept_pref}

As noted above there is a {\it constraint} not to exceed any professor's mandated teaching workload. Ideally, the total teaching units assigned to a professor will match their mandated workload. However, we found that imposing such a constraint makes our scheduling problem intractable. Moreover, while it is not permissible to exceed the mandated teaching workload, it is permissible (though not preferable) for the total assigned teaching units to be less than that workload. Thus, the algorithm should work to satisfy the {\it preference} that total assigned teaching units match the mandated workload while also imposing a {\it constraint} that they not exceed the mandated workload. 

Another departmental preference involves ``associated'' lecture and laboratory sections. Some courses have both a lecture and laboratory component. A typical lecture section has 48 students, while a laboratory section has 24 students. Thus, a lecture section typically has two associated laboratory sections. In Table~\ref{classes_offered} the lecture section PHYS-122-01 is associated with laboratory sections PHYS-122-02 and 122-03\cite{large lecture-lab sections}. For pedagogical reasons, it is preferable that associated lecture and laboratories have the same professor. Thus, the algorithm should work to satisfy that {\it preference}, by assigning for example PHYS-122 sections 01, 02, 03 to the same professor. This is not however, a {\it constraint}.

\subsection{Faculty Preferences}\label{fac_pref}


From our experiences, faculty members have very strong preferences for their teaching schedule. These preferences include which course(s) they are assigned, how many different courses they are assigned (the number of course preparations), and the time of day that they teach. Faculty members will view the quality of their teaching schedule through the lens of these preferences. We found perceived inequities in scheduling to come from two primary areas. The first is from professors who request (or demand) a preferred teaching schedule in advance of the actual scheduling process, and the second in is the arbitrary serial assignment approach typical of manual scheduling\cite{manual}. Both can lead to widespread, chronic dissatisfaction. 

One way to avoid such inequities would be to employ an automated computational approach that assigns classes while completely ignoring faculty preferences. This would, however, inevitably lead to widespread dissatisfaction, albeit of an equitable nature. Our goal was to develop an efficient, computer-automated scheduling process that attempts to meet faculty preferences fairly and transparently\cite{full_rnd}.

We did not find it practical to incorporate all types of preferences an individual faculty may have, but did identify the five most common preferences from a survey we deployed. In no particular order, they are:



\begin{enumerate}

\item {Not to teach 8am classes.} The earliest classes in our department begin at 8am, which can be an undesirable time slot for some. 

\item {Teach in the 1st or 2nd half of the day.} Some professors prefer morning or afternoon classes, so a preference is allowed for desiring classes that avoid the portion of the day they do not desire.  Here the first half of the day is considered 8am-1pm, and the second half 1pm-6pm.

\item {Teach preferred classes.} Some professors have particular classes they prefer to teach.

\item {Have minimal time-gaps between classes.} Professors generally prefer minimal daily time-gaps between their classes.

\item {Teach a minimal number of different courses.}  Some professors prefer to teach a minimal number of different classes in a given term.

\end{enumerate}

As with the departmental preferences, the algorithm should work (but is not constrained) to satisfy faculty preferences. Prior to assigning classes, faculty members are given the opportunity to specify their preferences by selecting a weighting (between 0.0 and 1.0) to each of the five preferences, indicating how important that particular preference is to them. 

For example, if a professor is ambivalent about teaching at 8am, they would assign a weighting of 0.0. If not teaching at 8am is of critical importance, they would assign it weighting of 1.0.  In an effort to ensure fairness, the sum of a professor's five weightings must equal 1.0.  A sample table of two professors' preferences is shown in Table \ref{preference table}.

\begin{table*}[t]
\centering
\begin{tabular}{c|c|c|c}
\toprule
Professor & Preference Type & Preference Selection & Preference Weighting \\
\midrule
Prof$_1$ & No 8am classes  & No & 0.4\\
Prof$_1$ & 1st or 2nd half of the day  & 1st half & 0.2\\
Prof$_1$ & Preferred classes & PHYS 121, PHYS 142 & 0.1\\
Prof$_1$ & Minimal time-gaps between classes & Yes & 0.2\\
Prof$_1$ & Minimal number of different courses & Yes & 0.1\\
$\vdots$ & $\vdots$ & $\vdots$ & $\vdots$ \\
Prof$_{29}$ & No 8am classes  & Yes & 0.3\\
Prof$_{29}$ & 1st or 2nd half of the day  & 2nd half & 0\\
Prof$_{29}$& Preferred classes & PHYS 143 & 0.3\\
Prof$_{29}$ & Minimal time-gaps between classes & Yes & 0.1\\
Prof$_{29}$ & Minimal number of different courses & Yes & 0.3\\
$\vdots$ & $\vdots$ & $\vdots$ & $\vdots$ \\
\hline
\end{tabular}
\caption{Examples of professor's preferences and preference weightings}\label{preference table}
\end{table*}


\subsection{Quantifying local ``fitness": a measure of preference satisfaction for a professor's teaching schedule}\label{constraints}

As will be seen in the next sections, the task of the GA is to produce a workable department schedule (professors assigned to classes) that satisfies ``hard'' constraints (no time conflicts and no exceeding of mandated teaching loads) while also maximally satisfying departmental and professor preferences (or ``soft'' constraints) described in the above two sections. To do so, one must devise a means of quantifying preference satisfaction. 

To this end, we define a local numerical ``fitness'' number for each professor's schedule. The better the satisfaction of preferences, the smaller that local fitness will be. The departmental, or ``global'', numerical fitness is the sum of the local fitnesses, and the goal of the GA is to minimize the global fitness.

The numerical fitness $f_p$ of a schedule for professor $p$ is a sum of seven $\Delta$ components, two that measure satisfaction of the departmental preferences (Section~\ref{dept_pref}), and five that measure satisfaction of that professor's faculty preferences (Section~\ref{fac_pref}). 

The components are described in the Appendix, and demonstrates how a single numerical fitness $f_p$ can be generated for each professor, $p$, and a department-wide fitness, $F$, can be found from $F=\sum_{p} f_p$.  The GA described here works to minimize $F$.

\section{The Genetic Algorithm: Framework}

\subsection{Assigning the classes: the chromosome encoding}

The GA implementation that will work to minimize the department-wide fitness, $F$, begins with a a binary string (or ``chromosome'') that must be constructed to drive the search for an optimal schedule.  Here, $N$ professors need placement into $M$ classes.  To begin, both professors and classes are enumerated with integers, so professor $p$ will be $0\le p \le N-1$ and class $c$ will be $0\le c\le M-1$.  In the binary string, chunks of $8$-bits are used to adequately contain both of these quantities.  

\begin{figure}[h]
\centering
\includegraphics[width=\columnwidth]{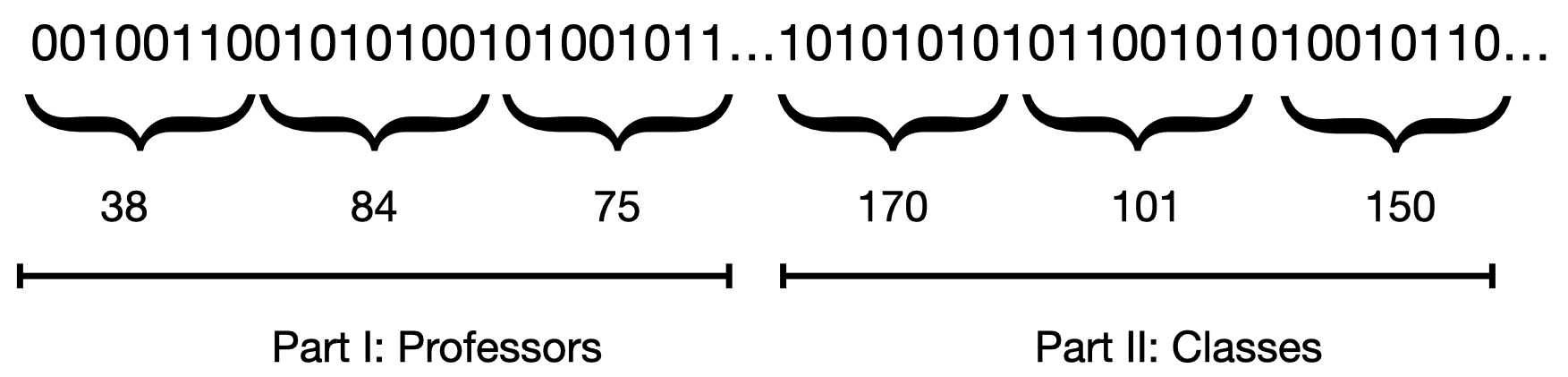}
\caption{The binary encoding or ``chromosome'' used in this work.  It consists of two concatenated parts, binary enumerations for the professors and classes. We use a byte size of $8$-bits. Here, professor $38$ will be paired with class $170$, professor $84$ with $101$ and so on. }
\label{binary_encoding}
\end{figure}

The chromosomes used here will be the concatenation of two parts, as shown in Figure~\ref{binary_encoding}.  Part 1 will be $8N$ bits to represent the professors, and Part 2 will be $8M$ bits to represent the classes.  In use, the professor in the first (decimal equivalent) $8$-bit chunk of Part 1 will be paired with the class in the first (decimal equivalent) $8$-bit chunk of Part 2. The professor in the second $8$-bit chunk of Part 1 with the class in the second $8$-bits chunk of Part 2, and so on. In Figure~\ref{binary_encoding}, professor $p={38}$ will be paired with class $c={170}$,  $p={84}$ with $c={101}$, and so on. 

Our actual chromosome is 2,000 bits in length, which we found adequate for $N=50$ and $M=200$\cite{2000bits}. It is acknowledged that when initialized with random bits,  Parts 1 and 2 may very well contain multiple references to the same professors and classes. We choose to let the GA grapple with this\cite{diff_encodings}.

Interpreting the chromosomes in this way handles {\sl assigning} of classes to professors, and fits naturally into the chromosome idea needed by a GA.

\subsection{Assessing the classes: the chromosome fitness}

With the chromosome encoding established, assessing a given schedule is now addressed.  This will be a numerical value, computable from the schedule represented by a chromosome. This work is designed so that a fitness of zero is a ``perfect'' schedule, thus this algorithm seeks to {\sl minimize} the fitness for a population of chromosomes.

In assigning classes to a professor given the chromosome pairings, a few checks are always in place that will summarily reject a pairing proposed by the chromosome.  These are: 1) When a pairing contains a class that has already been assigned, 2) if a class assignment would cause a professor to have more units than they are contractually  obligated to teach, or 3) if a class conflicts in time with another class that a professor is already assigned.  

With such checks in place, a schedule produced from a chromosome can be considered ``clean,'' meaning that basic scheduling requirements are being met: each class is only assigned to one professor, contractual unit violations will not occur, and no one will have any time conflicts in their schedules. Such a clean schedule could in fact even be deployed, and it would minimally work.

The numerical fitness is derived from seven components, all computed by parsing and analyzing the schedule that a chromosome represents.  The seven parts, initially summarized in Section~\ref{constraints} described above and in the Appendix, with $f_p$ being the fitness of the schedule proposed for professor $p$.

\subsection{Adapting the class assignments: the optimization}

The GA is now set off in search of an optimal schedule as follows, using the standard implementation discussed both in Refs.~\citenum{koza} and~\citenum{michalewicz}.  Briefly, $100$ initial chromosomes are created, each out of a uniformly random sequence of 1s and 0s. These $100$ chromosomes serve as the initial generation. 

\begin{figure}[h]
\centering
\includegraphics[width=\columnwidth]{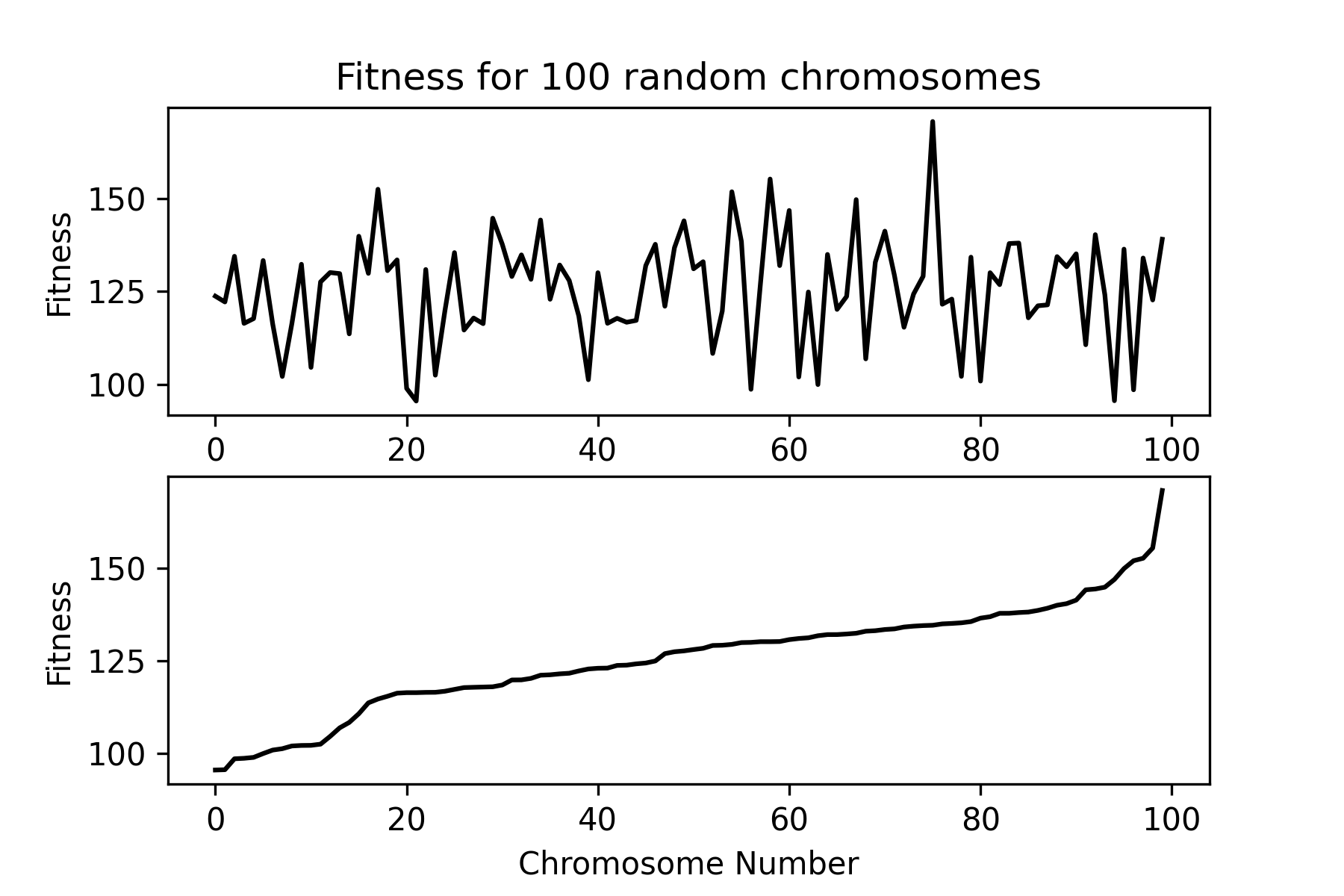}
\caption{Fitness values for an initial generation of random chromosomes (top: unsorted, bottom: sorted and relabeled).}
\label{rnd_pop_fitness}
\end{figure}

In Step 1 of adapting the schedule, the fitness function is computed for each chromosome in the generation. Next the chromosomes are sorted according to their fitness (and relabeled), as shown in Figure~\ref{rnd_pop_fitness}.

In Step 2, the sorted population is used as its own probability density to drive a ``roulette wheel'' selection method\cite{koza,michalewicz} to select two individual chromosomes with a selection probability that is inverse to their fitness. (So chromosomes with a lower fitness have a higher probability of being selected.)  The roulette wheel selection is done by first finding the maximum fitness value of the chromosomes in a generation, $f_{\textrm{max}}$.  A chromosome $k$ is selected when

\begin{equation}
\sum_{i=0}^k \left[f_{\textrm{max}}-g_i\right] > r\sum_{i=0}^{100} \left[f_{max}-g_i\right],
\end{equation}
for some minimum $k$, where $r$ is a uniform random number on $0\le r \le 1$ and $g_i$ is the fitness of chromosome $i$ in the sorted population.

With a pair of ``more fit'' chromosomes selected, two actions occur.  The first is a crossover. Here, a random index point is selected between $0$ and the length of a chromosome.  All bits between this point and the end of the chromosome are swapped between the pair.  Second, there is mutation where the bits of the two new chromosomes (that resulted from the crossover) are visited one-by-one, and bit-flipped ($1\rightarrow 0$ and $0\rightarrow 1$) with a probability of $0.01$\cite{norvig}.  

The two new chromosomes are saved as the first two in a new generation. This process is repeated until the new generation contains $100$ (new) chromosomes.  This new generation replaces the original generation and Step 1 is repeated on the new generation, resulting in yet another generation.  In the algorithm, the probability to a crossover was set to $25\%$.

We repeatedly monitor the population and look for an algorithm-wide minimum fitness for the entire department ($F$ in Eqn.~\ref{dept_fitness}) amongst the chromosomes in the evolving generations. When one occurs, we (greedily) grab it and hold the chromosome separately.   The algorithm terminates when a lower algorithm-wide minimum is no longer found, which typically occurs after $400$ generations.  The chromosome having the algorithm-wide minimum will be the best schedule produced. 

This process reflects the final desired action of a scheduling algorithm, continually {\sl adapting} the schedule proposed by the chromosomes (which once again is a natural part of a GA), to make it more optimal.

\section{The Genetic Algorithm: Interpretability}

With the GA framework in place, we are now ready to run the algorithm and examine its work in finding an optimal schedule.  In this case, it was run for a term having $N=52$ professors needing to be placed into $M=155$ classes.  The flexibility to break class associations was not allowed (so $\Delta_{assoc}=0$ as described in the Appendix).  

\subsection{Evolution of the generations}
\begin{figure}[h]
\centering
\includegraphics[width=\columnwidth]{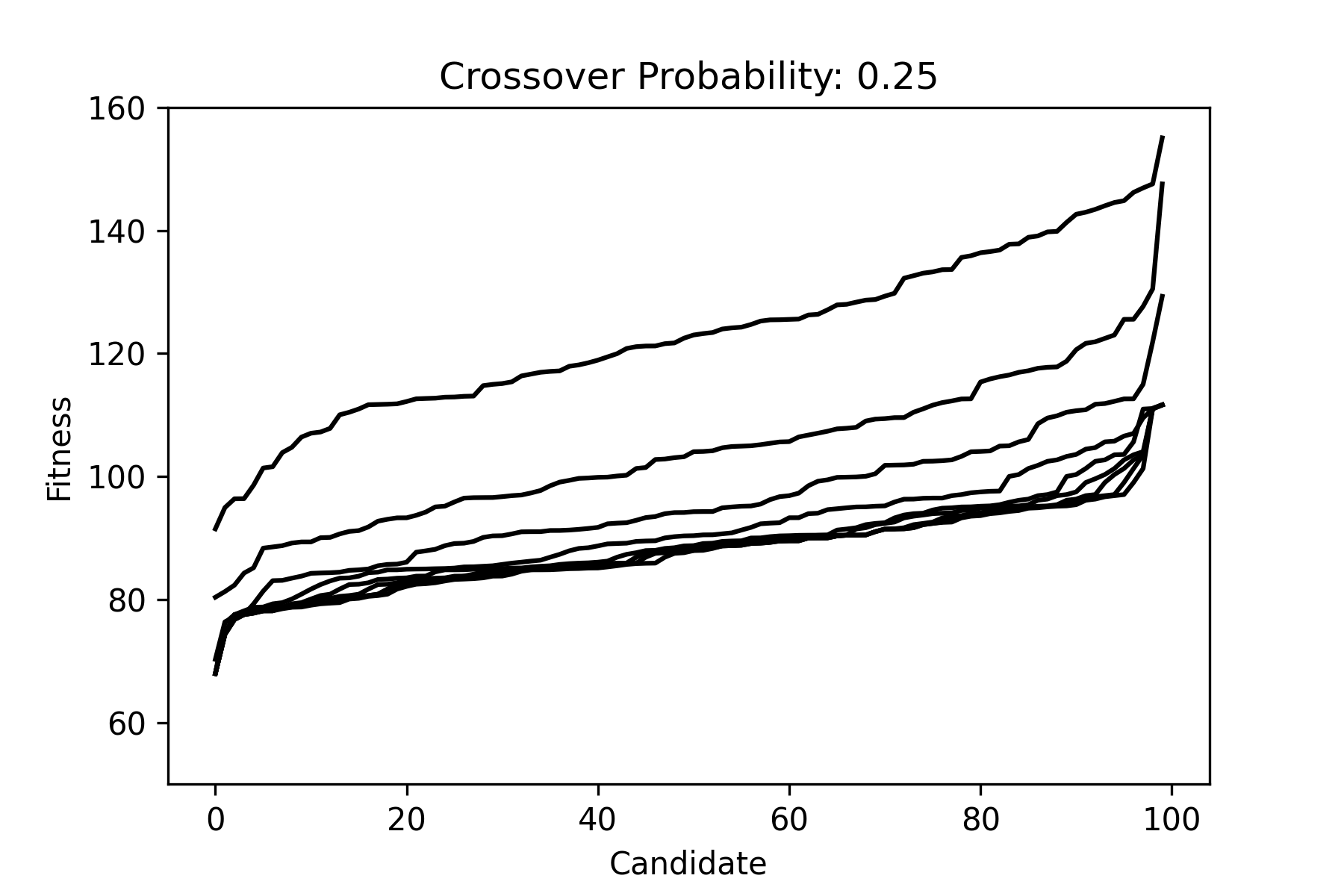}
\caption{Evolution of the algorithm's global minimum chromosome over select generations.}
\label{xo_025}
\end{figure}

Figure~\ref{xo_025} shows the evolutions of the (sorted) chromosome fitness for $9$ generations (of the $400$ run) that produced a schedule with a new algorithm-wide minimum fitness.  Generation $0$ is the topmost curve, followed by a clear downward trend that shows convergence of the algorithm, as evidenced by the overlapping (darkened area) curves at the bottom of the group.  The GA indeed appears to be working as the fitness of subsequent generations decreases.

\begin{figure}[h]
\centering
\includegraphics[width=\columnwidth]{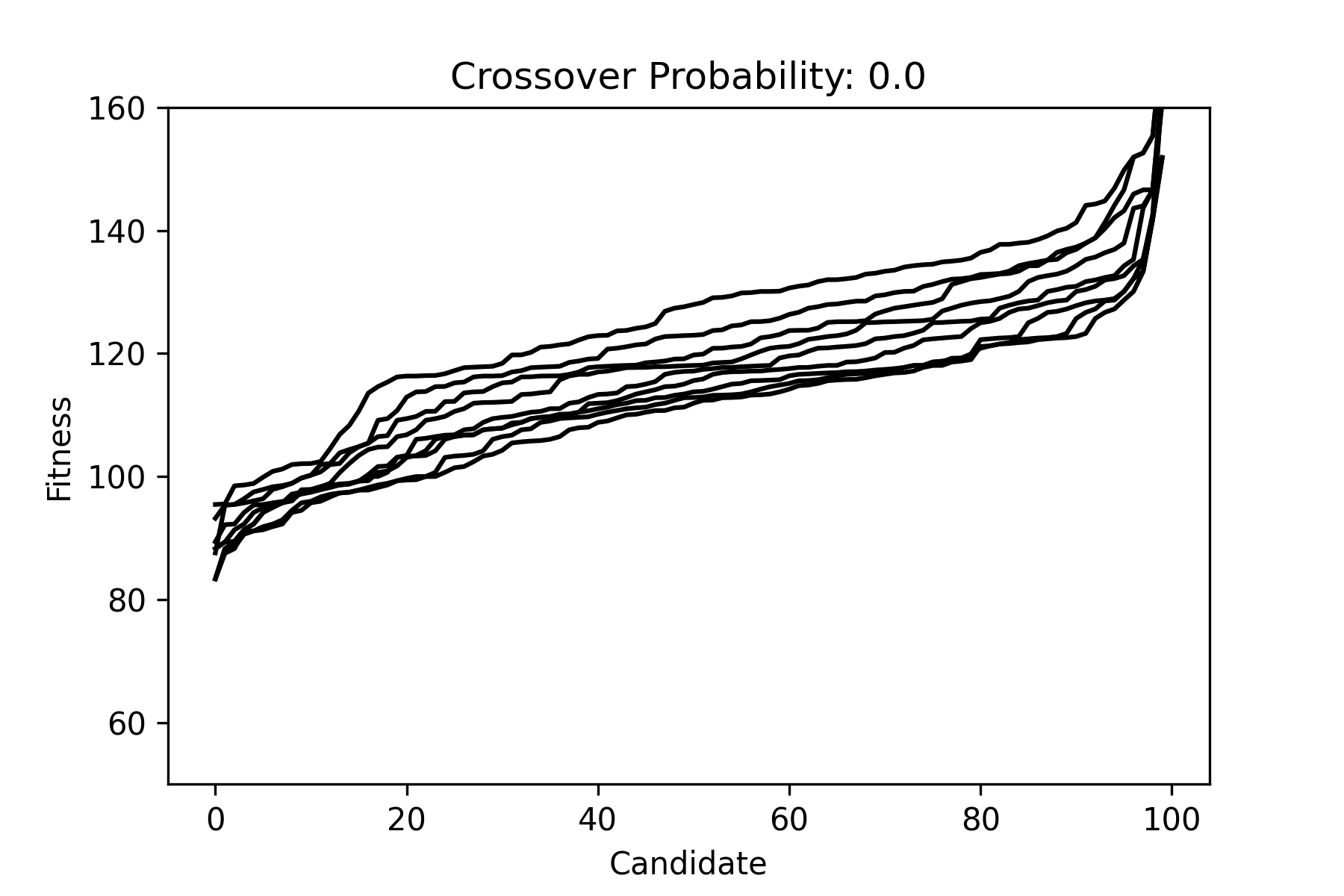}
\caption{Evolution of the algorithm's global minimum chromosome over select generations, with the crossover operator turned off.}
\label{x0_00}
\end{figure}

Being somewhat skeptical of the crossover operation singlehandedly carrying the entire optimization, the GA was run with a crossover probability of zero (disallowing crossovers).  Mutations were still allowed, and the evolving generations were again captured as shown in Figure~\ref{x0_00}. Only a small decline in the fitness is seen, as compared to Figure~\ref{xo_025}. This is due to the roulette selection still used to select chromosomes for the next generation (after mutation).  Indeed the crossover operation is driving the optimization, and this shows how random numbers alone will not lead to any systematic optimization.

\subsection{Fitness and assigned class count}
\begin{figure}[h]
\centering
\includegraphics[width=\columnwidth]{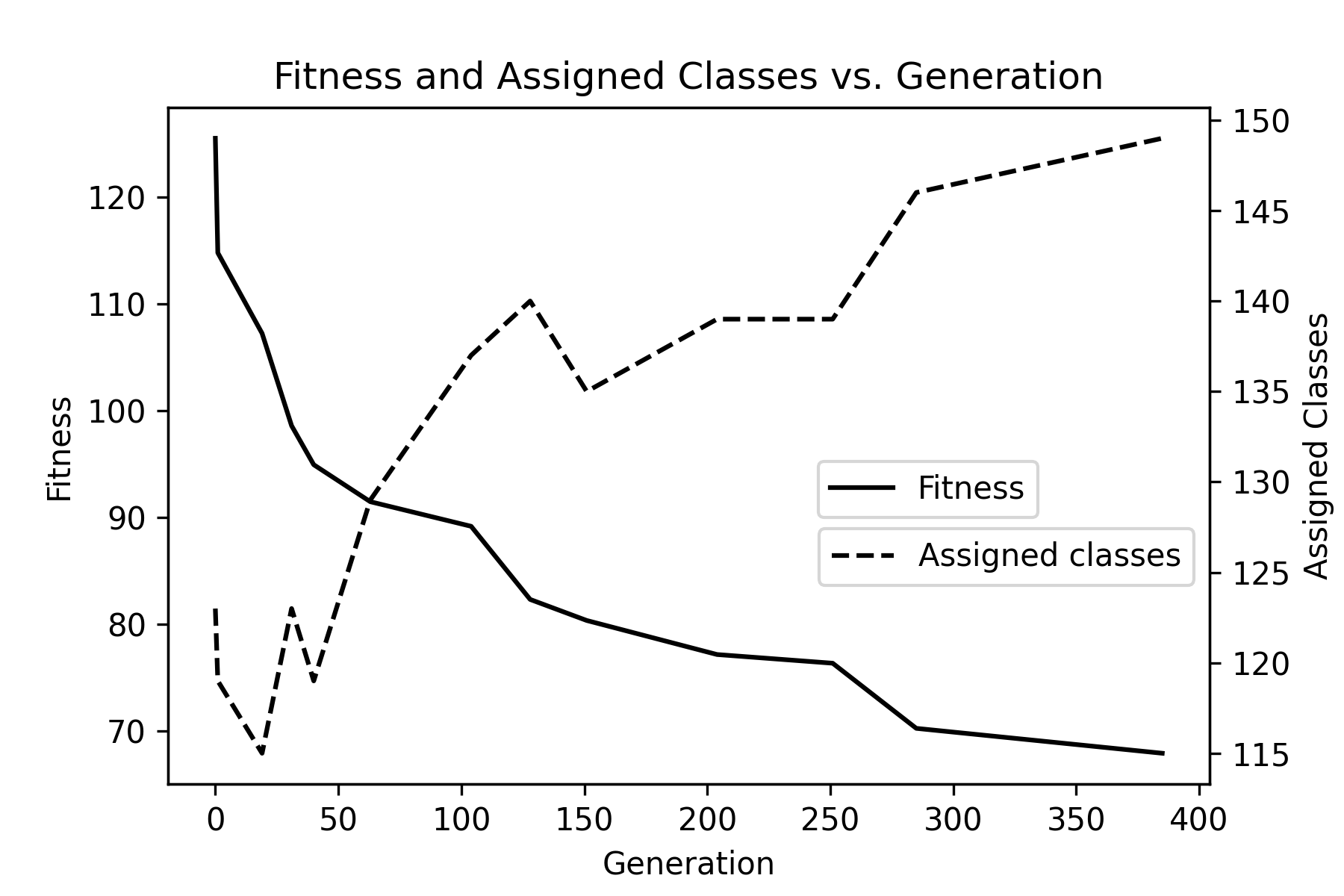}
\caption{The minimum fitness (left axis) found in a given generation and number of assigned classes (right axis), vs. generation number.}
\label{fitness_vs_epoch}
\end{figure}

An overall look at the performance of the GA is shown in Figure~\ref{fitness_vs_epoch}, where the solid curve represents the algorithm-wide minimums found and the dotted curve the number of classes assigned. At generation 0, which represents the best schedule from a generation of completely random chromosomes, only about $80$ classes were assigned, with a fitness of $125$.  As the algorithm proceeds, we see two encouraging trends by generation 400: 1) the fitness drops to by a factor of $1.8$ to $67$ and 2) the number of assigned classes grows to $149$. This is further evidence that the algorithm is working as the fitness is decreased and a schedule is produced with $96\%$ of the classes assigned.

\subsection{Fitness for each professor}
\begin{figure}[h]
\centering
\includegraphics[width=\columnwidth]{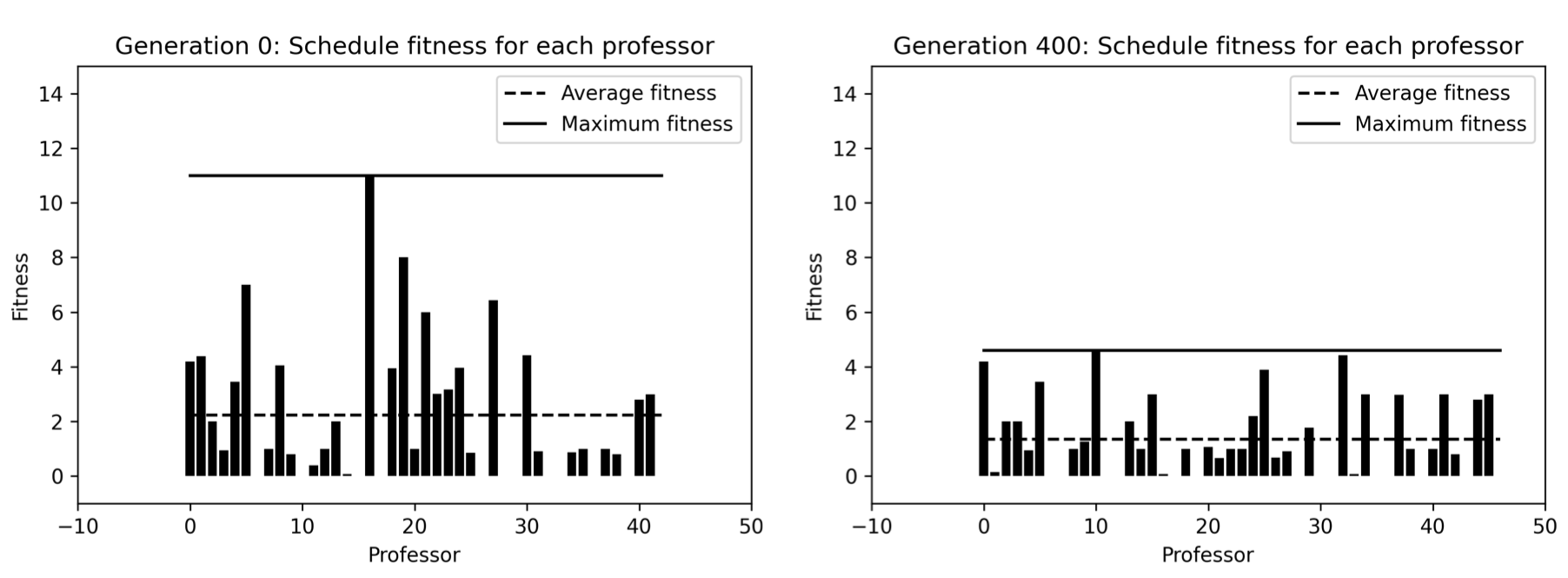}
\caption{The fitness of each professor at generations 0 (fully random) and 400 (optimized to convergence). Empty bars
are for professors who were prescheduled (by hand) and were not given any classes by the GA, or those who did not lodge any personal preferences so all of their weights in Equation~\ref{fitness_func} are zero.}
\label{share_the_pain}
\end{figure}

Next we examine how the fitness for each professor evolved during the optimization, which is shown in Figure~\ref{share_the_pain},  where each bar represents the fitness of the schedule proposed for a given professor.  In both plots, the solid line indicates the maximum fitness across all professors and the dotted line the average fitness.  The left plot is generation $0$ (fully random) and the right is generation $400$, when the algorithm has converged. This shows still more evidence that the algorithm is working, since both the maximum and average bars decrease from generation $0$ to $400$. As a purely visual guide, the solid line in the right figure appears to bound a tighter envelope around the fitness values, indicating that not only has the fitness for each professor dropped, but they have all dropped together.  This is a critical demonstration of fairness in the schedule, showing a {\sl distribution in fairness}\cite{muhlenthaler2013} in the end result.

\subsection{Explainability}

\begin{figure}[h]
\centering
\includegraphics[width=\columnwidth]{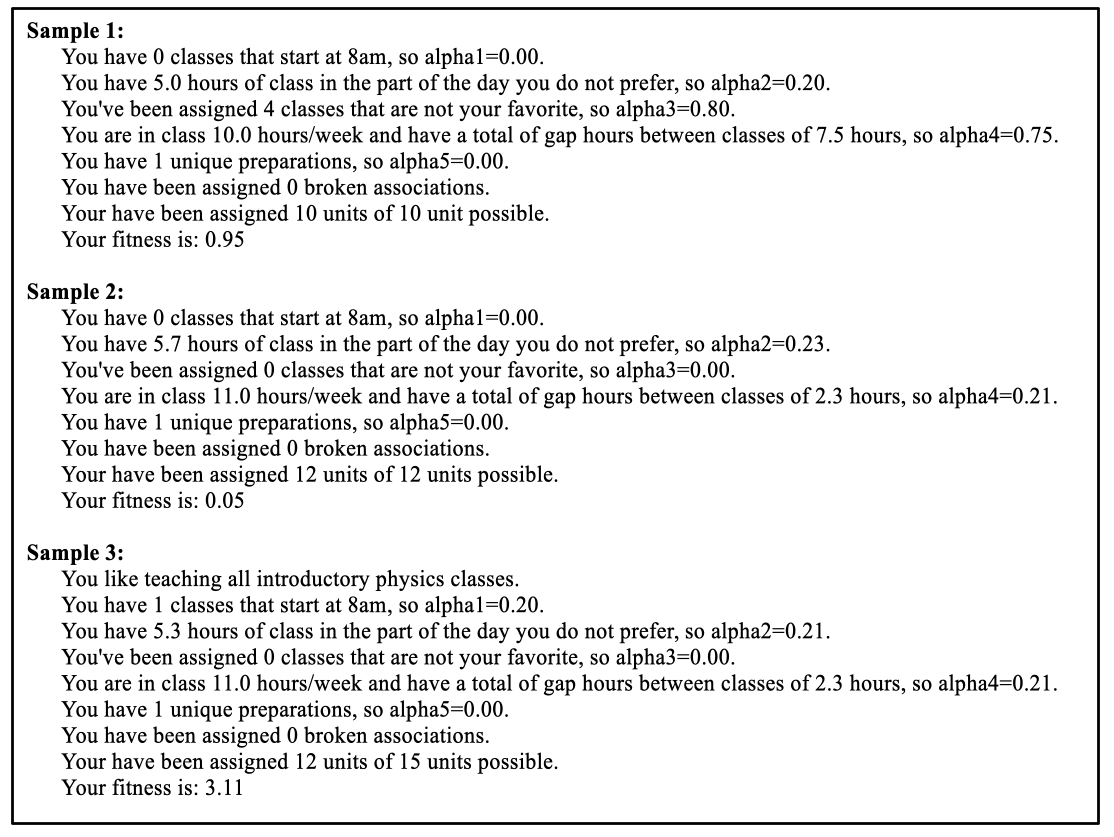}
\caption{Text output presented to each professor, as compiled by the function that computes the fitness of professor's schedule}
\label{explain}
\end{figure}

Explainability is an important outcome of computer-generated solutions\cite{explain,xu}, as it is known to increase the trust of the end-user in a computationally generated solution.  In this work, text-based explanations are included in the code that computes the fitness for the schedule of a given professor. Since the fitness function contains $7$ terms as shown in Equation~\ref{fitness_func},  the explanation generally contains $7$ statements.  Three samples are shown in Figure~\ref{explain}. 

Thus, in addition to one's schedule fitness and how it compares with others, each professor is also offered an explanation of how their fitness was computed. This may help them to understand why their schedule resulted as such, and how their preferences were taken into account.

\subsection{Performance summary}
We find all the interpretability analysis shown here to be strong evidence that a {\sl fair} schedule is being produced.  It is clear that the fitness started high for each professor, and ended up low as likely possible when the GA converges. We attribute this to the class assignment action of the GA: it assigns, assesses, and adjusts classes for everyone {\sl in parallel}, a drastically different approach to the {\sl serial} method typically done with manual scheduling.  The explainable fitness results should help each professor understand the origin of their schedule and how their teaching efforts are being used to help the department as a whole.

\section{Using Prolog for further optimizations}

We are generally pleased with the ability of the GA to produce our schedules, which are mostly acceptable directly from the algorithm. They still go through a thorough human-level review, which typically involves some minor edits prior to being released. Most edits involve placing the few remaining classes and/or breaking up long  sequences of back-to-back classes that may have appeared.

In the manual reviews, a few common inconveniences are noted to regularly occur, but are not global enough to warrant adding them to the fitness function of the GA. Instead, the Prolog language\cite{clocksin} is used to search for these oddities in a completed schedule.

First, the schedule is dumped to a Prolog source file as shown in Figure~\ref{prolog_schedule_facts}. As a quick Prolog tutorial\cite{clocksin}, lowercase functions are considered ``facts'' in Prolog, so here are facts called {\tt class} that describes the entire schedule (to Prolog). The fact {\tt class} holds the identification of the professor, the assigned class name, the day of the week the class runs, the start and end times and the room.

\begin{figure}[h]
{\footnotesize
\begin{verbatim}
class(prof1,astr-102-01,mtwr,[17,10,18,00],room_180_0101).
class(prof2,phys-202-01,tr,[09,10,11,00],room_180_0272).
class(prof2,phys-202-02,tr,[14,10,16,00],room_180_0272).
class(prof2,phys-428-02,m,[12,10,15,00],room_180_0634).
class(prof2,phys-428-03,f,[12,10,15,00],room_180_0634).
class(prof3,phys-142-43,mwf,[10,10,11,00],room_180_0262).
class(prof3,phys-142-44,mwf,[11,10,12,00],room_180_0262).
class(prof4,phys-142-21,mwf,[13,10,14,00],room_053_0206).
class(prof4,phys-142-22,m,[15,10,18,00],room_180_0269).
class(prof4,phys-142-23,t,[08,10,11,00],room_180_0269).
class(prof5,geol-417-01,r,[15,10,18,00],room_180_0233).
class(prof5,geol-203-01,tr,[09,40,11,00],room_053_0201).
class(prof5,geol-203-02,t,[12,10,13,00],room_180_0233).
...
...
\end{verbatim}
}
\caption{Prolog facts describing a given schedule, as dumped by the GA code. Prolog rules may be applied against these facts to search for scheduling quirks (see text).}
\label{prolog_schedule_facts}
\end{figure}

With the scheduling facts known, logical queries can now be constructed to see if Prolog can verify if the facts support the queries or not.

As a test, we first check for any day/time conflicts among the classes (Prolog variables start with an uppercase letter) in a clause called {\tt conflict}:

{\footnotesize
\begin{verbatim}
conflict :-
        class(Prof,Class1,Days1,Times1,Room1),
        class(Prof,Class2,Days2,Times2,Room2),
        Class1 \= Class2,
        Days1 = Days2,
        Times1 = Times2,
        format("Conflict between ~w and ~w\n",Class1,Class2).
\end{verbatim}
}

Here two classes are chosen {\tt Class1} and {\tt Class1} that have the same professor {\tt Prof}, as well as the days, times, and rooms of each class. Then {\tt Class1} and {\tt Class2} are confirmed to be different, and checked if they are offered at the same days and times. If so, a warning message is displayed.  Running {\tt conflict} in SWI-Prolog\cite{swi} always returns {\tt false}, as such conflicts are forbidden by the GA assignment algorithm described above.

An actual situation that commonly occurs is as follows. Suppose a professor has classes in back-to-back time slots (i.e. 12:10pm-1:00pm and 1:10pm-2:00pm) but in different rooms, as shown in Tables~\ref{can_swap1} and ~\ref{can_swap2}. It would be nice if the professor could remain in the same room for the sake of class logistics (such as the set up of audio/visuals, tangibles, lesson aides, demonstrations, etc.).   A swap can be arranged if a scheduling pattern in two different rooms occurs as shown in these tables.

\begin{table}[h]
\centering
\begin{tabular}{c|c|c|c|c|c}
\toprule
 Time & M & T & W & TH & F\\
\midrule
$\vdots$ & $\vdots$ & $\vdots$ & $\vdots$ & $\vdots$ \\
11-12  & \\
12-1  & Prof1 & Prof1 & Prof1 & Prof1\\
1-2    & Prof2 & Prof2 & Prof2 & Prof2\\
2-3 & \\
$\vdots$ & $\vdots$ & $\vdots$ & $\vdots$ & $\vdots$ \\
\hline
\end{tabular}
\caption{Room 1}
\label{can_swap1}
\end{table}

\begin{table}[h]
\centering
\begin{tabular}{c|c|c|c|c|c}
\toprule
 Time & M & T & W & TH & F\\
\midrule
$\vdots$ & $\vdots$ & $\vdots$ & $\vdots$ & $\vdots$ \\
11-12  & \\
12-1  & \\
1-2    & Prof1 & Prof1 & Prof1 & Prof1 & Prof1\\
2-3 & \\
$\vdots$ & $\vdots$ & $\vdots$ & $\vdots$ & $\vdots$ \\
\hline
\end{tabular}
\caption{Room 2}
\label{can_swap2}
\end{table}
Here, Prof1's class in Room 2, and Prof2's class in Room 1 should be swapped, allowing Prof1 to stay in same room. Such possible room-swap possibilities are difficult to spot manually in a completed schedule. For this, the clause {\tt should$\_$swap} was constructed as:

{\footnotesize
\begin{verbatim}
should_swap :-
    class(Prof1,C1,Days,Times1,Room1),
    class(Prof1,C2,Days,Times2,Room2),
    
    hour_before_or_after(Times1,Times2),

    Room1 \= Room2,

    class(Prof2,C,Days,Times,Room),

    Prof2 \= Prof1,
    
    (Times = Times1 ; Times = Times2),

    format("~w is ~w ~w in ~w (~w)\n",
    			[C1,Days,Times1,Room1,Prof1]),
    format("~w is ~w ~w in ~w (~w)\n",
    			[C2,Days,Times2,Room2,Prof1]),
    format("~w is ~w ~w in ~w (~w)\n",
    			[C,Days,Times,Room,Prof2]).
\end{verbatim}
}

This clause looks for two classes {\tt C1} and {\tt C2}, both assigned to professor 1 ({\tt Prof1}). Then we check that the two classes are back-to-back using the helper logic in {\tt hour$\_$before$\_$or$\_$after} (not shown) and that the rooms of the two classes are different.

Next another class {\tt C} by professor 2 ({\tt Prof2}) is searched for, that has the same day-to-day meeting pattern ({\tt Days}). If the professors are different and either of the class times of the professor 2's classes are the same as either of professor 1's classes, then a simple swap between the rooms can be done.  Often, 10 or so possible swaps are found in a given schedule, as the one shown here:

{\footnotesize
\begin{verbatim}
phys-123-1 is mwf [10,10,11,0] in room_053_0202 (prof1)
phys-142-18 is mwf [11,10,12,0] in room_053_0201 (prof1)
phys-142-32 is mwf [10,10,11,0] in room_053_0201 (prof2)
\end{verbatim}
}

In this case, the {\tt phys-142-32} class assigned to {\tt prof2} should be swapped with the {\tt phys-123-1} class assigned to {\tt prof1}. This enables {\tt prof1} to remain in room {\tt 53-201} for their two classes.

Other clauses examine if broken associations can possibly be recombined into a single professor's schedule or if there are possibilities of reducing the number of course preparations with simple course swaps between pairs of professors. To date, we have developed ${\sim}6$ such searches that can lead to small changes that we feel can make a give professor's schedule a bit better.  Prolog appears to be a useful tool for implementing complex searches in this scheduling environment.

\section{Conclusions}
We have reported on the use of a genetic algorithm, utilizing the standard crossover and mutation operations, to assign professors to classes in a large physics department. 

We have shown how a GA naturally satisfies the ``three As'' of scheduling, which are difficult to grapple with when doing scheduling manually. 

The first is in {\sl assigning} classes to professors, which comes directly from an instantaneous interpretation of the chromosomes. This was the most gratifying to outcome, as this work allowed us to elevate our technique out of a manual scheduling mode, which often led to the ability to make higher-level optimizations or quickly pivot the entire schedule as issues arose.  

The second is in {\sl assessing} the quality of a given schedule. This was done by computing a fitness function based on some mandatory and some optional preferences, with optional preferences coming directly from the professors themselves. This allows us to demonstrate a {\sl distribution of fairness} (see Ref.~\citenum{muhlenthaler2013} and Fig.~\ref{share_the_pain}) along with some explainability (see Fig.~\ref{explain}) with each scheduling iteration.

The last was {\sl adapting} the schedule, as done by the crossover and mutation operations. These operations propose wholly different schedules to consider, which were incrementally more optimal than previous ones, since the operations were performed by mixing characteristics of the most optimal schedules pulled from a previous generation.

The algorithm could typically make over $90\%$ of the needed course assignments before converging.  At this point, typically only minimal handwork is needed in order to fully complete a schedule. Interpreting the algorithm, we confirm a lowering of the fitness relative to the initial random schedule seed, for subsequent generations produced by the algorithm.  We include explainable text into the fitness calculations, for the benefit of the end-user and use Prolog to further refine the schedule before releasing it.  

Finally, a locally produced schedule such as this one must be input into a larger university-wide system for administrative needs and inclusion into the student course selection portal.  Like scheduling itself, this can also can be an arduous and error-prone task.  To complement the GA, we also developed a Python script to automatically input the schedule into the university system using a browser automation tool called Selenium\cite{selenium}, directly from a CSV file produced by the GA codebase\cite{bensky_selenium}.

\section{Acknowledgements}

We thank M. Moelter for proofreading this paper.  TB is grateful to KS for his work in revamping the components of the fitness function. TB also sends a special thank you to the Evolutionary Computation and Machine Learning group at Victoria University in Wellington, New Zealand, for their kind and stimulating hospitality from January to April of 2024, where much of this paper was written.

\appendix

\section{Appendix: Fitness components}

Earlier in the paper, three actions were listed as needed of the automated scheduler: assigning, assessing, and adapting. This appendix shows how the development of a fitness function, as needed of a GA, handles {\sl assessing} a given schedule.

\subsection{Departmental preference: assigned schedule meets professor's mandated workload}

The first component is $\Delta^{(units)}_p$, which is equal to the mandated workload of professor (measured in teaching units) minus the teaching units assigned to professor $p$. For example, if professor $p$ has a mandated workload of 13 teaching units, but is assigned a schedule with 11 teaching units, then  $\Delta^{(units)}_p = 2$. Since the algorithm is constrained to only consider schedules that do not exceed the mandated workload,  $\Delta^{(units)}_p \geq 0$ for each professor. 

The algorithm should work to minimize $\Delta^{(units)}_p$, hopefully even find a $\Delta^{(units)}_p=0$ for as many professors as possible.

\subsection{Departmental preference: schedule has associated lecture and laboratory sections}

As discussed above, it is preferable that associated of lecture and laboratory sections are taught by the same professor. For example, PHYS 122-01 (lecture) and PHYS 122-02, PHYS 122-03 (laboratories) should be assigned to the same professor.   Experience shows it is usually impossible to complete an entire schedule without ``breaking'' such associations, but it is important to  minimize such breaks\cite{manual_breaks}.

This fitness component $\Delta^{(assoc)}_p$ is thus a measure of how many broken associations are part of the schedule assigned to professor $p$. For example, if professor $p$ is assigned PHYS 122-02, but not PHYS 122-01 and PHYS 122-03, then $\Delta^{(assoc)}_p=2$. If another professor $q$ is assigned PHYS 122-01 and PHYS 122-03, but not PHYS 122-02, then $\Delta^{(assoc)}_q=1$.

The algorithm should work to minimize $\Delta^{(assoc)}_p$, meaning breaks in course associations are minimized for as many professors as possible.
 
 \subsection{Faculty preference 1: Prefer not to teach 8am classes}
 
Like the departmental preferences, we seek a quantity $\Delta^{(8am)}_p$ that measures how well a professor's preference {\sl not} to teach at 8am is met, while taking into account the relative importance of that preference to the professor (the preference weighting). 

We thus define $\Delta^{(8am)}_p=\alpha^{(8am)}_p w^{(8am)}_p$, where $\alpha^{(8am)}_p$ is a measure of preference satisfaction, and $w^{(8am)}_p$ is the relative importance weighting of that preference. For example, in Table \ref{preference table}, $w^{(8am)}_1=0.4$ and $w^{(8am)}_{29}=0.3$. We define $\alpha^{(8am)}_p = D_p/5.0$, where $D_p$ is the number of days that professor $p$ has a class at 8:00am. The denominator of $5$ is a worst-case normalizer of having 8:00am classes $5$ days a week. As an example, if professor $p$ has an 8:00am class each day of the week, then $\alpha^{(8am)}_p=1$. 

If professor 1 in Table~\ref{preference table} has an 8am class 3 days/week, then $\Delta^{(8am)}_1=0.24$, whereas for professor 29, $\Delta^{(8am)}_{29}=0.18$ for the same number of 8am classes. 

The algorithm should work to minimize $\Delta^{(8am)}_p$, meaning as few professors as possible who do not like teaching 8am courses will get assigned a class at that time.

\subsection{Faculty preference 2: Prefer to teach in the 1st or 2nd half of the day}

The next contribution to a professor's fitness is $\Delta^{(half)}_p = \alpha^{(half)}_p w^{(half)}_p$, with  $\alpha^{(half)}_p$ measuring preference satisfaction, and $w^{(half)}_p$ preference weighting. 

This accounts for the preference of professor $p$ having classes in the first or second half of the day so we define $\alpha^{(half)}_p=H^{NP}_p/H^{total}_p$, where $H^{NP}_p=$ the number of weekly hours professor $p$ spends teaching in their {\it non-preferred} (NP) half of the day. $H^{total}_p$ (hours) is the total number of weekly teaching hours, and is a worst-case normalizer of having 100\% classes in the non-preferred half of the day.

For example, professor 1 from Table~\ref{preference table} is assigned 12 hours/week of classes, and prefers to teach in first half of day. Their assigned courses meet in the first half of the day, except for a lecture that meets three days/week, 2-3pm. This corresponds to three hours per week teaching in the wrong half of the day. Thus, $\alpha^{(half)}_1=3/12=0.25$. Since $w^{(half)}_1=0.2$, the corresponding $\Delta^{(half)}_1=0.05$.

The algorithm should work to minimize $\Delta^{(half)}_p$, meaning as few professors as possible will have classes in the half of the day they do not prefer.

\subsection{Faculty preference 3: Prefer to teach favorite classes}

The next contribution is $\Delta^{(fav)}_p = \alpha^{(fav)}_p w^{(fav)}_p$ which is a preference about which of the six non-major courses are favored by professor $p$. We define $\alpha^{(fav)}_p={\tilde F}_p/(6-F_p)$, where ${\tilde F}_p$ is the number of non-favorite courses assigned to professor $p$, and $F_p$ the number of their favorite courses they indicated as a preference. For example, PHYS 121 and PHYS 142 are the classes favored by professor 1 from table \ref{preference table}, so $F_1=2$. If they are assigned PHYS 121 and/or PHYS 142 then ${\tilde F}_1=0$ and $\alpha^{(fav)}_p=0$ meaning that their preference has been perfectly satisfied. However, if professor 1 is assigned PHYS 122, 123, 141, and 142, none of which are their favorite classes, then ${\tilde F}_1=4$ and $\alpha^{(fav)}_1=1$, the maximum possible value, which reflects the worst-case scenario. Taking into account the importance of this preference for professor 1, $\Delta^{(fav)}_1=0.1$.

The algorithm should work to minimize $\Delta^{(fav)}_p$, meaning as few professors will teach classes they do not prefer.

\subsection{Faculty preference 4: Prefer to have minimal time-gaps between classes.}

The next contribution is $\Delta^{(gap)}_p = \alpha^{(gap)}_p w^{(gap)}_p$ which is a preference to minimize time gaps between meeting times of courses. We define $\alpha^{(gap)}_p=H_p/35$ where $H_p$ is the total number of gap-hours between classes in a given week in the schedule of professor $p$. Consider the worst-case scenario in which professor $p$ has a class 8am-9am and another 5pm-6pm, giving a $7$ hour gap, five days per week.  For this, $H_p=7\times 5=35$, giving $\alpha^{(gap)}_p=1$ (hence the $35$ in the definition). Smaller gaps would give a smaller values of $\alpha^{(gap)}_p$.

The algorithm should work to minimize $\Delta^{(gap)}_p$, meaning professors all have a minimal gap time between their classes.

\subsection{Faculty preference 5: Prefer to teach a minimal number of different courses.}

The final contribution is $\Delta^{(prep)}_p = \alpha^{(prep)}_p w^{(prep)}_p$ which is a preference to minimize the number of different non-major course preparations assigned to professor $p$ in a give term. We define $\alpha^{(prep)}_p = (N_p-1)/5$ where $N_p$ is the number of different courses assigned to professor $p$. In the best case scenario, professor $p$ has only one course preparation and $\alpha^{(prep)}_p=0$. In the worst-case scenario, professor $p$ is assigned all six non-major courses and $\alpha^{(prep)}_p=1$.

The algorithm should work to minimize $\Delta^{(pre)}_p$, meaning professors have as few different class preparations as possible.

 
\subsection{The fitness equation}

One can imagine more $\alpha$-style parameters customized for other kinds of scheduling work. With those given, the local fitness $f_p$ for professor $p$'s schedule is defined to be

\begin{align}
f_p=\Delta^{(units)}_p+\Delta^{(assoc)}_p + \Delta^{(8am)}_p + \Delta^{(half)}_p+ \notag \\
\Delta^{(fav)}_p + \Delta^{(gap)}_p+ \Delta^{(prep)}_p
\label{fitness_func}
\end{align}
with the $\Delta$ parameters defined above. The global fitness for the entire department is then

\begin{equation}
F=\sum_p f_p,
\label{dept_fitness}
\end{equation}
which is what the GA works to minimize.

\bibliography{sn-bibliography}

\begin{thebibliography}{99}
\bibitem{manual}  One can imagine a list of unassigned classes (with their teaching units) appearing next to a list of professors with empty schedules. In our department, there are $\sim{}50$ professors and $\sim{}200$ class sections. The  challenge is to pull unassigned classes, and place them with a given professor until a given professor's schedule meets their contractural obligation. Even tangibles used to track this process can be challenging, which must allow some ``back and forth'' in use. This has included colored sticky notes, cut slips of paper with a name or class on it, sliding magnetic labels on a whiteboard, clear, erasable acetate sheets or (most predominantly) colorful Excel spreadsheets, some quite sophisticated with or without use of the built-in scripting language.
\bibitem{non teaching workload} Faculty members with a teaching workload less than 15 units, typically have other non-instruction responsibilities such as committee service and/or grant-related scholarship.
\bibitem{room_optimize} Optimizing class placement into available rooms is another area in need of automation and optimization.
\bibitem{Table Totals} The sum of teaching workload units in Table~\ref{prof_capacity} should equal that of Table~\ref{classes_offered}.
\bibitem{large lecture-lab sections} There are a small number of 72 student lecture section that have three associated laboratory sections.
\bibitem{full_rnd} We implemented code to do scheduling in this manner at the beginning of this effort, and the results were very obviously terrible.
\bibitem{manual_breaks} Breaking associated classes is one of the great downfalls of manual scheduling, as it is every tempting to do so in order to fill in small workload-unit gaps in one's schedule and get all classes assigned, particularly as a scheduling deadline is looming.
\bibitem{diff_encodings} Different chromosome encodings were examined. In one encoding, the length of each part shown in Figure~\ref{binary_encoding} was expanded to an integer multiple of $N$ and $M$, to allow for the possibility that multiple unique professor orderings might be found in Part I.  In another encoding, $p$ and $c$ were pulled from the chromosome as {\sl adjacent} binary chunks in the chromosome.  Neither showed any profound (dis)advantage on the algorithm.
\end{thebibliography}

\end{document}